\documentclass[conference]{IEEEtran}
\IEEEoverridecommandlockouts

\usepackage{amsmath}
\usepackage{amssymb}
\usepackage{amsfonts}
\usepackage{amsthm}
\usepackage{algorithm}
\usepackage{algpseudocode}
\usepackage{graphicx}
\usepackage{textcomp}
\usepackage{xcolor}
\usepackage{booktabs}
\usepackage[numbers,sort&compress]{natbib}
\usepackage{array}
\usepackage{mathtools}
\usepackage{hyperref}

\newcommand{\li}[1]{{#1}}
\begin{document}

\title{TD3 with Reverse KL Regularizer for Offline Reinforcement Learning from Mixed Datasets
\thanks{This work is conducted at Microsoft Research Asia.}}

\author{
\IEEEauthorblockN{Yuanying Cai$^{1}$, Chuheng Zhang$^{1,2}$, Li Zhao$^{2,*}$\thanks{* Corresponding Author}, Wei Shen$^3$, Xuyun Zhang$^4$,\\ Lei Song$^2$, Jiang Bian$^2$, Tao Qin$^2$, Tieyan Liu$^2$}
\IEEEauthorblockA{$^1$Tsinghua University, Beijing, China\\
$^2$Microsoft Research Asia, Beijing, China\\
$^3$Independent Researcher\\
$^4$Macquarie University, Sydney, Australia\\
{cai-yy16@mails.tsinghua.edu.cn}, zhangchuheng123@live.com, shenwei0917@126.com, \\ xuyun.zhang@mq.edu.au, \{lizo, lei.song, jiang.bian, taoqin, tyliu\}@microsoft.com}}

\maketitle

\begin{abstract}
We consider an offline reinforcement learning (RL) setting where the agent need to learn from a dataset collected by rolling out multiple behavior policies.
There are two challenges for this setting: 
1) The optimal trade-off between optimizing the RL signal and the behavior cloning (BC) signal changes on different states due to the variation of the action coverage induced by different behavior policies.
Previous methods fail to handle this by only controlling the global trade-off.
2) For a given state, the action distribution generated by different behavior policies may have multiple modes.
The BC regularizers in many previous methods are mean-seeking, resulting in policies that select out-of-distribution (OOD) actions in the middle of the modes.
In this paper, we address both challenges by using adaptively weighted reverse Kullback-Leibler (KL) divergence as the BC regularizer based on the TD3 algorithm.
Our method not only trades off the RL and BC signals with per-state weights (i.e., strong BC regularization on the states with narrow action coverage, and vice versa) but also avoids selecting OOD actions thanks to the mode-seeking property of reverse KL.
Empirically, our algorithm can outperform existing offline RL algorithms in the MuJoCo locomotion tasks with the standard D4RL datasets as well as the mixed datasets that combine the standard datasets.
\end{abstract}

\section{Introduction}
\label{sec:introduction}

In recent years, offline reinforcement learning~(RL) achieves great success on many real-world applications where online data collection is risky and expensive such as robotics \citep{mandlekar2020gti,kahn2021badgr}, healthcare \citep{johnson2016mimic}, advertising \citep{liao2022cross}, and dialogue systems \citep{jaques2019way}.
In offline RL, the agent aims to learn a good policy from previously collected dataset without further interaction with the environment. 
Although (online) off-policy RL methods are applicable to the offline setting, directly using these methods can result in suboptimal performance due to the distribution shift problem~\citep{lange2012batch}: 
The distribution induced by the learned policy is different from the distribution over the offline dataset.
Consequently, we cannot estimate the values of out-of-distribution state-action pairs accurately, and thus the policy may take overestimated out-of-distribution actions leading to suboptimal performance.
This common pathway through which online RL algorithms can fail in the offline RL setting~\cite{fujimoto2019off} motivates later studies on offline RL.

To alleviate the distribution shift problem, the key for the offline RL setting is to learn a reasonably conservative policy that can avoid visiting out-of-distribution state-action pairs while optimizing the performance of the policy.
Policy-based offline RL methods usually adopt various techniques to constrain the learned policy to be close to the behavior policy (i.e., the policy used to collect the offline dataset)~\citep{fujimoto2019off,wu2019behavior,kumar2019stabilizing,siegel2020keep,kostrikov2021offline,fujimoto2021minimalist}.
In other words, there are two signals in the training process: 
The RL signal that trains the agent to maximize the cumulative reward and the behavioral cloning~(BC) signal that constrains the learned policy to be close to the behavior policy. 

\begin{figure}[t]
   \centering
   \includegraphics[width=0.7\columnwidth]{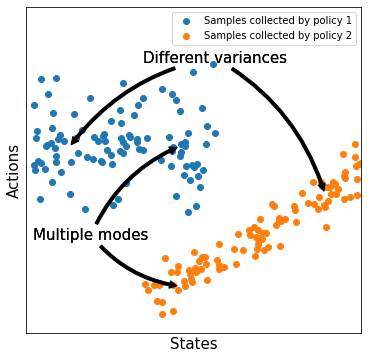}
   \caption{
   Illustration on the state-action pairs in an offline dataset collected by two behavior policies.
   The variance of the action samples changes across different states, which calls for trading off the RL and BC signals adaptively on different states.
   The action samples can present multiple modes on a same state where the learned policy should avoid taking the out-of-distribution actions in the middle of the modes.
   }
\label{fig:example}
\end{figure}

The success of offline RL also depends on the quality and the size of the dataset.
A common scenario in the industry is that we are provided with a large dataset with mixed samples 
collected using different behavior policies.
However, different from \li{the case where data samples are} collected with a single behavior policy,
such mixed datasets bring new challenges that prevent us from learning a reasonably conservative policy.
We find that existing offline RL methods are not designed for mixed datasets and perform poorly in our later experiments with mixed datasets.

As an example, we show an offline dataset collected by two different policies in Figure \ref{fig:example}.
For ease of presentation, we consider one-dimensional states and actions.
In this example, the two policies visit different parts of the state space 
\li{and produce different action distributions with different variances.}
Specifically, action samples generated by the first policy (e.g., a random policy) have a larger variance than those generated by the second policy (e.g., an expert policy).
The challenges arise from the following two distributional properties of the state-action samples in the mixed dataset:

First, we observe that the variance of the action samples varies on different states,
\li{which motivates} us to adaptively 
trade off between the RL and BC signals.
For the states with small action coverage (cf. the states covered only by the second policy), it is better to use a strong BC signal since we have little knowledge on the effect of other actions.
In contrast, for the states with a large action coverage (cf. the states visited by the first policy), we can rely more on the RL signal to choose a good action from \li{the distribution of sampled actions}. 
However, although previous methods consider the trade-off between the RL and BC signals, they do not adjust the trade-off on different states adaptively.
This may lead to suboptimal performance when the dataset is collected by behavior policies with different levels of stochasticity.

Second, the distribution of the action samples in certain states is \li{multimodal}. 
Previous methods that use the mean-squared-error \citep{fujimoto2021minimalist} or the Kullback-Leibler (KL) divergence \citep{wu2019behavior} as the BC regularizer are mean-seeking, i.e., encouraging the policy to take the mean action \li{of} the action samples.
However, when the action samples present multiple modes (cf. the states visited by both policies in Figure \ref{fig:example}), such regularization encourages the policy to take the actions in the middle of the modes that may be out-of-distribution.

To \li{address} these challenges, we propose a simple yet effective method that uses adaptively weighted reverse KL divergence between the learned policy and the \li{behavior action distributions} 
as the BC regularizer. 
On one hand, the mode-seeking property of the reverse KL divergence (i.e., encouraging the policy to select the actions from one of the modes) can prevent the policy from selecting out-of-distribution actions when the behavior action distribution is highly multi-modal.
On the other hand, we weight the BC regularization adaptively for a given state with a transformation of the aleatoric uncertainty (or the standard deviation) of the action samples on this state. 
Then, we combine the BC regularization using reverse KL divergence with one of the state-of-the-art offline RL algorithms TD3+BC \citep{fujimoto2021minimalist} resulting in our algorithm called TD3+RKL.
We compare TD3+RKL with several strong baselines on the MuJoCo locomotion tasks with the standard D4RL dataset \citep{fu2020d4rl}.
Empirically, we find that TD3+RKL not only outperforms these baselines when learning from datasets collected using a single behavior policy but also achieves significantly better performance on mixed datasets that are collected using different policies.

Our contributions are summarized as follows: 

\begin{itemize}
    \item We consider a special offline RL setting that requires the agent to learn from mixed dataset collected by multiple distinct behavior policies. 
    We summarize the two challenges for this setting: different variances of the action samples on different states and the multi-modality of the behavioral action distribution on certain states.
    \item To face these challenges, we propose TD3+RKL that uses adaptively weighted reverse KL divergence as the BC regularizer.
    We show that this simple technique not only adjusts for the trade-off between the RL and BC signals on different states adaptively but also avoids selecting out-of-distribution actions when the behavior action distribution is highly multi-modal.
    \item We empirically show that TD3+RKL outperforms the previous methods on most of the offline RL tasks using the D4RL dataset and achieves the best performance when the dataset is collected by a mixture of behavior policies.
\end{itemize}

\section{Related Work}
\label{sec:related_work}

In this paper, we propose a new form of behavior cloning (BC) regularization for the offline RL setting that learns from datasets collected by a mixture of behavior policies.
To handle different variances of the behavior action samples on different states, we weight the BC regularizer according to the aleatoric uncertainty of the action samples.
In this section, we provide a brief survey on offline RL from mixed datasets, different BC regularizers in offline RL, and using uncertainty in offline RL.

\textbf{Offline RL from mixed datasets.}
Although most offline RL formulation accepts the dataset collected from multiple behavior sources, few papers focus on learning from such mixed datasets.
However, this setting is very common in real-world problems.
Many previous methods rely on estimating the behavior policy $\pi_\beta(\cdot | s)$ with a uni-modal Gaussian model \citep[e.g.,][]{fujimoto2019off,kumar2019stabilizing,wu2019behavior,jaques2019way,siegel2020keep,simao2019safe}.
However, a uni-modal Gaussian model may fail to estimate the highly multi-modal action distribution accurately and can result in a policy that selects out-of-distribution actions~\citep{levine2020offline}.
To deal with multi-modal action distribution, several previous methods circumvent explicit behavior policy estimation by using samples to approximate the behavior distribution.
For example, \citet{kumar2020conservative} propose CQL that tries to increase the estimated Q values on behavior state-action samples while decrease those collected by the target policy.
\citet{peng2019advantage} and \citet{nair2020accelerating} present advantage-weighted forms of behavior cloning that maximizes the weighted log-probability that the target policy can generate the behavior data.
Similar to these methods, we derive a BC regularizer that can be calculated based on samples and avoid estimating the behavior policy.
Moreover, we focus on the changing variance of the behavior action distribution on different states and design a mode-seeking BC regularizer to deal with this challenge.

\textbf{Behavior cloning in RL.}
Behavior cloning (BC) signals/regularizers are used in both online and offline RL settings.
In online RL, BC signals are used to accelerate the learning process \citep{hester2018deep,nair2020accelerating}, encourage exploration \citep{nair2018overcoming,flet2021adversarially}, impose safety constraints \citep{malik2021inverse}, or overcome the sparse reward problem \citep{goecks2020integrating}.
In offline RL, previous policy-based methods incorporate various forms of BC signals, including the divergence regularization between the target and the behavior policies (e.g., KL divergence \citep{jaques2019way,fakoor2021continuous}, maximum mean discrepancy, \citep{kumar2019stabilizing}, or others \cite{wu2019behavior}), or direct behavior cloning regularizers \cite{wang2018exponentially,peng2019advantage,nair2020accelerating,nair2020awac}.
Our paper uses reverse KL divergence as the BC regularization in offline RL, which will be shown to address the challenges in offline RL from mixed datasets.
Several previous papers such as BRAC \citep{wu2019behavior} also use the reverse KL divergence.
However, minimizing the reverse KL divergence leads to an entropy maximization of the target policy and may result in an overly exploratory policy when using stochastic target policies as in these papers.
We overcome this limitation by learning a deterministic target policy.

\textbf{Using uncertainty in offline RL.} 
We use \emph{aleatoric uncertainty} (i.e., the inherent uncertainty in the dataset) to balance RL and BC signals on different states, whereas previous offline RL methods usually use \emph{epistemic uncertainty} (due to the lack of data)~\citep{kendall2018multi,kendall2017uncertainties} to detect out-of-distribution state-action pairs.
Specifically, these methods treat the state-action pairs with large epistemic uncertainty as out-of-distribution and encourage the learned policy to stay away from the OOD state-action pairs in both model-free methods~\citep{buckman2020importance,kumar2020conservative,liu2020provably,wu2021uncertainty,jin2021pessimism} and model-based methods~\citep{yu2020mopo,kidambi2020morel}.

\section{Preliminary and Background}
\label{sec:preliminary}

\begin{table*}
\caption{\normalfont Comparison of the behavior cloning regularization in the update rules of different policy constraint based methods.}
\label{tab:comp_rule}
\centering
\abovedisplayskip=0pt
\belowdisplayskip=0pt
\begin{tabular}{ m{2cm} m{14cm} }
\hline
Algorithms & Update Rules \\
\hline
BCQ \citep{fujimoto2019off} & 
{\begin{flalign}
\text{maximize}_{\xi_\phi} \mathbb{E}_{(s,a)\sim\mathcal{D}}
\Big[ 
Q_\theta\big(s,a'+\xi_\phi(s,a')\big) \Big| a' \sim \hat{\pi}_b (\cdot | s)
\Big]
&&
\label{eq:BCQ}
\end{flalign}} 
\\
EMaQ \citep{ghasemipour2021emaq} & 
{\begin{flalign}
\text{arg max}_{a\sim\hat{\pi}_b(\cdot|s)}
\Big[ 
Q_\theta\big(s,a\big)
\Big],
\text{ and }
Q_\theta\big(s,a\big) \leftarrow 
r(s,a) + \gamma \mathbb{E}_{s'}
\left[\max_{a' \sim\hat{\pi}_b(\cdot|s')} Q_\theta(s',a')\right]
&&
\label{eq:EMaQ}
\end{flalign}} 
\\
BEAR \citep{kumar2019stabilizing} & 
{\begin{flalign}
\text{maximize}_{\pi_\phi} \mathbb{E}_{(s,a)\sim\mathcal{D}}
\Big[
Q_\theta\big(s,\pi_\phi(s)\big) 
- \lambda \text{MMD} \big(
\hat{\pi}_b(\cdot|s) \big|\big| \pi_\phi(\cdot|s)
\big)
\Big]
&&
\label{eq:BEAR}
\end{flalign}} 
\\
CDC \citep{fakoor2021continuous} & 
{\begin{flalign}
\text{maximize}_{\pi_\phi} 
\mathbb E_{(s,a)\sim\mathcal{D}}
\Big[
Q_\theta\big(s,\pi_\phi(s)\big) 
- \lambda \text{KL} \big(
\hat{\pi}_b(\cdot|s) \big|\big|\pi_\phi(\cdot|s) \big)
\Big] = 
\mathbb E_{(s,a)\sim\mathcal{D}}
\Big[
Q_\theta\big(s,\pi_\phi(s)\big) 
+ \lambda \log \pi_\phi(a|s) \big)
\Big]
&&
\label{eq:CDC}
\end{flalign}}
\\
BRAC \citep{wu2019behavior} & 
{\begin{flalign}
\text{maximize}_{\pi_\phi} \mathbb{E}_{(s,a)\sim\mathcal{D}}
\Big[
Q_\theta\big(s,\pi_\phi(s)\big) 
- \lambda \text{KL} \big( \pi_\phi(\cdot|s)\big|\big|\hat{\pi}_b(\cdot|s) \big)
\Big]
&&
\label{eq:BRAC}
\end{flalign}} 
\\
TD3+BC \citep{fujimoto2021minimalist} & 
{\begin{flalign}
\text{maximize}_{\mu_\phi}\mathbb{E}_{(s,a)\sim\mathcal{D}}
\Big[
Q_\theta\big(s,\mu_\phi(s)\big) - \lambda \big(\mu_\phi(s)-a\big)^2
\Big]
&&
\label{eq:TD3+BC}
\end{flalign}} 
\\
TD3+RKL (ours) & 
{\begin{flalign}
\text{maximize}_{\mu_\phi} \mathbb E_{(s,a)\sim\mathcal{D};\textcolor{cyan}{a_1,a_2\sim\mathcal{D}}}
\Bigg[
Q_\theta\big(s,\mu_\phi(s)\big) 
-
\textcolor{purple}{\lambda(s)}
\left(
\left(\mu_\phi(s) - a \right)^2 
- \textcolor{cyan}{\alpha \left(\mu_\phi(s) - \frac{a_1+a_2}{2} \right)^2}
\right)
\Bigg]
&&
\label{eq:TD3+RKL}
\end{flalign}} 
\\
\hline
\end{tabular}
\end{table*}

\subsection{Offline Reinforcement Learning}

We consider the discounted infinite-horizon Markov decision process (MDP) $\mathcal{M}=(\mathcal{S}, \mathcal{A}, P, r,\rho, \gamma)$, where $\mathcal{S}$ and $\mathcal{A}$ are the state space and action space respectively, $P: \mathcal{S}\times \mathcal{A}\to \Delta^{\mathcal{S}}$ is the transition dynamics, $r: \mathcal{S}\times \mathcal{A}\to \mathbb{R}$ is the reward function, $\rho$ is the initial state distribution, and $\gamma\in[0,1]$ is the discounted factor~\citep{sutton2018reinforcement}. 
Given a policy $\pi: \mathcal{S} \to \Delta^\mathcal{A}$, the return starting from the state-action pair on the $t$-th step $(s_t, a_t)$ is defined as the sum of the discounted rewards $R_t^\pi = \sum_{\tau=t}^\infty \gamma^{\tau-t} r(s_\tau, a_\tau)$ where $s_{t+1}, a_{t+1}, s_{t+2}, a_{t+2}, \cdots$ are collected by rolling out the policy $\pi$ starting from $(s_t, a_t)$.
The objective of online reinforcement learning (RL) is to learn a policy that maximizes the expected return $J(\pi)=\mathbb{E}_{s_0 \sim \rho, a_0 \sim \pi(\cdot|s_0)}[R_0^\pi]$. 
Given a state-action pair $(s,a)$ and a policy $\pi$, the Q function is defined as $Q^\pi(s,a)=\mathbb{E}[R_0^\pi|s_0=s, a_0=a]$ where the expectation is taken over all the possible trajectories starting from $(s,a)$.
The Q function is the fixed point of the following Bellman evaluation operation~\cite{fujimoto2018addressing}:
\begin{equation}
    \label{eq:bellman_eval}
    \mathcal{T}^\pi Q(s,a) := r(s,a)+\gamma\mathbb E_{s'\sim P(\cdot|s,a),a'\sim \pi(\cdot|s')}[Q(s',a')].
\end{equation}
Besides, the Q function of the optimal policy $\pi^*$ denoted as $Q^*:=Q^{\pi^*}$ is the fixed point of the following Bellman optimality operation: 
\begin{equation}
    \label{eq:bellman_op}
    \mathcal{T}^* Q(s,a) := r(s,a)+\gamma\mathbb E_{s'\sim P(\cdot|s,a)}\left[\max_{a'}Q(s',a')\right].
\end{equation}
For the offline RL setting, the objective is to learn a policy that maximizes the expected return with the provided offline dataset $\mathcal{D}=\{(s_t, a_t, r_t, s'_t)\}_{t=1}^N$ instead of interactions with the environment~\cite{levine2020offline}. 
Furthermore, we consider the setting where the dataset is collected by a mixture of \emph{behavior policies} denoted as $\pi_b$.

\subsection{Offline RL with Policy-Based Constraints}

As previously introduced, the key for offline RL is to control distribution shift by learning a conservative policy that can avoid visiting out-of-distribution state-action pairs.
One category of the methods impose constraints on the learned policies with behavior cloning (BC) regularizers to encourage the learned policy to be close to the behavior policy.
We summarize the BC regularizers in several popular existing methods in Table \ref{tab:comp_rule}.
Previous offline RL methods either model the target policy as the estimated behavior policy $\hat{\pi}_b$ plus a perturbation 
or learn parameterized policies directly.

The representatives of the first category include BCQ~\citep{fujimoto2019off} and EMaQ \citep{ghasemipour2021emaq}.
BCQ models the target distribution as $\hat{\pi}_b(\cdot|s) + \xi_\phi(s)$ where $\hat{\pi}_b$ is the estimated behavior policy and $\xi_\phi$ is a parameterized perturbation network.
EMaQ simplifies BCQ by removing the perturbation network at the cost of more computational costs at the testing time (i.e., sampling from estimated behavior policy multiple times to select a best action).
As shown in Eq. \eqref{eq:BCQ} and Eq. \eqref{eq:EMaQ} in the table, BCQ and EMaQ can evaluate and optimize the Q values only on the state-action pairs near the behavior state-action samples from the dataset with the help of such modeling.
These methods construct the target policy on top of the estimated behavior policy and therefore rely on the quality of the estimated behavior policy.
However, when the dataset is generated by multiple distinct behavior policies with complex patterns in the state-action distribution, it is hard to estimate the mixture of behavior policies accurately.
Therefore, the success of these methods largely depends on a careful design on the generative model used to approximate the behavior policies~\citep{ghasemipour2021emaq}.

The second category of methods learn a deterministic policy (denoted as $\mu_\phi: \mathcal{S} \to \mathcal{A}$) or a stochastic policy (denoted as $\pi_\phi: \mathcal{S} \to \Delta^\mathcal{A}$) directly.
For example, as shown in Eq. \eqref{eq:BEAR}-\eqref{eq:BRAC}, BEAR \citep{kumar2019stabilizing}, CDC \citep{fakoor2021continuous} and BRAC \citep{wu2019behavior} constrain the policy using the maximum mean discrepancy (MMD), the forward KL divergence and the reverse KL divergence with the pre-estimated cloned policy $\hat{\pi}_b$ respectively\footnote{
Although the author claims that they use a reverse KL,
CDC actually uses a forward KL regularizer in which the latter term is the learnable distribution according to the definition in, for example, \citep{malinin2019reverse,chan2021greedification}.
}.
The benefit of using forward KL divergence is that is does not require sampling from the target policy $\pi_\phi$.
However, we will later show that the forward KL divergence is mean-seeking, which means that it cannot prevent the target policy from selecting out-of-distribution actions between multiple modes of the action samples.
BRAC uses the reverse KL divergence which is mode-seeking an can avoid selecting out-of-distribution actions.
\li{However, BRAC models the target policy as a parameterized stochastic policy, and consequently minimizing the reverse KL divergence induces a term that maximizes the entropy of the target policy.
This term incentivizes an overly exploratory policy which is not suitable for offline RL.
Moreover, BRAC still requires an pre-estimated behavior policy $\hat{\pi}_b$.}
TD3+BC \citep{fujimoto2021minimalist} proposes a simple method that is free from modeling a complex target policy (e.g., using the perturbation network or a stochastic policy) and estimating the behavior policy and achieves impressive performance. 
However, the MSE regularization used in TD3+BC is also mean-seeking and may result in suboptimal policies when learning from mixed datasets.
Later, we will introduce our method that \li{inherits} 
the simplicity of TD3+BC but can handle mixed datasets.

\subsection{KL Divergence in Offline RL}

\begin{figure}[tb]
   \centering
   \includegraphics[width=\columnwidth]{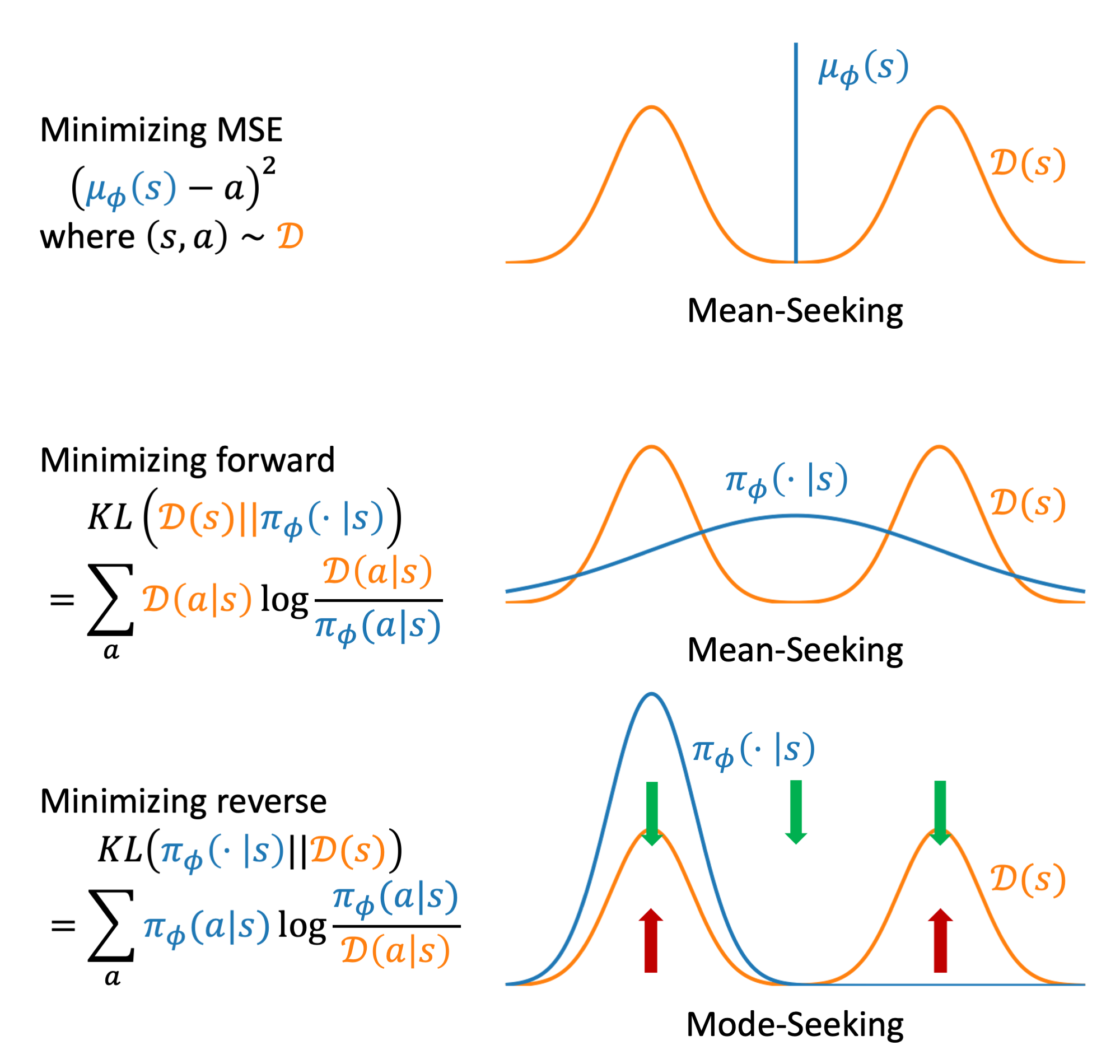}
   \caption{
    Comparison between different behavior cloning regularizers.
    The red and green arrows correspond to the first term and the second term in Eq. \eqref{eq:contrastive} respectively.
   }
\label{fig:reverse_KL}
\end{figure}

Consider two distributions over a space $\mathcal{X}$: a data distribution denoted as $p(x)$ and a parameterized distribution $q_\theta(x)$ to approximate the data distribution for some $x\in\mathcal{X}$.
The \emph{forward} and \emph{reverse} Kullback-Leibler (KL) divergence \citep{kullback1951information,anzai2012pattern} are defined as follows:
\begin{equation}
    \label{eq:forward_KL}
    \text{Forward KL: }
    KL(p||q_\theta)=\sum_x p(x)\log\frac{p(x)}{q_\theta(x)}
\end{equation}
\begin{equation}
    \label{eq:reverse_KL}
    \text{Reverse KL: }
    KL(q_\theta||p)=\sum_x q_\theta(x)\log\frac{q_\theta(x)}{p(x)}
\end{equation}

In offline RL, although all BC regularizers try to match the target policy with the data distribution, using different regularizers captures different properties of the data distribution.

We compare the effect of the MSE, reverse KL and forward KL regularizer in Fig.~\ref{fig:reverse_KL}.
In our example, the mixture of behavior policies generates an action selection probability with two modes denoted as $\mathcal{D}(s)$ on the state $s$.
As shown in Fig.~\ref{fig:reverse_KL}(a), MSE motivates a deterministic policy that outputs an out-of-distribution action in the middle of the two modes.
In Fig.~\ref{fig:reverse_KL}(b), we show that the learned policy $\pi_\phi(\cdot|s)$ under the forward KL regularizer covers the support of the data distribution $\mathcal{D}(s)$ and samples out-of-distribution actions with large probability.
This results from the formulation of forward KL where $\pi_\phi(\cdot|s)$ appears in the denominator which encourages the learned policy to sample the actions with nonzero probability in the region where $\mathcal{D}(a|s) > 0$.
In the middle of the two modes, the difference between $\pi_\phi(\cdot|s)$ and $\mathcal{D}(a|s)$ is ignored by forward KL since the weight $\mathcal{D}(a|s)$ vanishes in this area.
Therefore, the forward KL is also mean-seeking.
In Figure \ref{fig:reverse_KL}(c), we show that the learned policy $\pi_\phi(\cdot|s)$ under the reverse KL regularizer captures one of the modes in $\mathcal{D}(\cdot|s)$. 
Such mode-seeking effect is what we need since it helps us to avoid sampling out-of-distribution actions.

In offline RL, previous methods usually use the forward KL as BC signals to encourage the learned policy to be closed to the behavior policy~\citep{levine2020offline}. 
As we have mentioned before, the policy learned by minimizing the forward KL divergence covers the whole support of the behavior policy. 
Hence, it is not suitable \li{to use the forward KL regularizer to} learn a deterministic policy \li{on} the standard D4RL dataset where the behavior policies used to collect the dataset are stochastic policies. 




\section{Methodology}
\label{sec:method}

In this section, we introduce our algorithm TD3+RKL that regularizes the policy using the reverse KL divergence with samples collected by behavior policies.
Specifically, we consider learning a deterministic policy and derive the mode-seeking regularizer used in our algorithm starting from the reverse KL divergence formulation.
Moreover, to adaptively balance the RL and BC signals, we weight the BC regularization on different states according to the aleatoric uncertainty on the states.
At last, we present our practical algorithm.

\subsection{Mode-Seeking Regularizer}

As previously introduced, we need a mode-seeking regularizer to prevent the policy from selecting out-of-distribution actions.
Here, we first consider the following learning objective with a BC regularizer using the reverse KL divergence:
\begin{equation}
\label{eq:objective}
\text{max}_{\pi_\phi} \mathbb{E}_{s\sim\mathcal{D}}
\Big[
Q_\theta\big(s,\pi_\phi(s)\big) 
- \lambda(s) \text{KL} \big( \pi_\phi(s)\big|\big|\hat{\pi}_b(s) \big)
\Big],
\end{equation}
where the weight $\lambda$ is adaptive depending on different states.
This adaptive weight can be used to balance the RL and BC signals on different states and the detailed formulation will be introduced later.
We can rewrite the reverse KL divergence as follows:
\begin{equation}
\begin{aligned}
& \text{KL} \big( \pi_\phi(s)\big|\big|\hat{\pi}_b(s) \big) \\
= & \sum_{a\in\mathcal{A}} \pi_\phi(a|s) \log \frac{\pi_\phi(a|s)}{\hat{\pi}_b(a|s)} \\
= & - \mathcal{H}\big(\pi_\phi(a|s)\big)
 - \sum_{a\in\mathcal{A}} \pi_\phi(a|s) \log \hat{\pi}_b(a|s). \\
\end{aligned}
\end{equation}

We can see that minimizing the reverse KL divergence leads to maximization on the entropy of the target policy.
This term encourages exploration which may be useful for online RL but should be avoided in offline RL since an exploratory policy makes it easy for the target policy to select out-of-distribution actions.
To avoid an overly exploratory target policy, we model the target policy as a Gaussian policy with a fixed standard deviation $\sigma$, i.e., $\pi_\phi(s) = \mathcal{N}(\mu_\phi(s), \sigma)$.
Notice that modeling the target policy with uni-modal Gaussian does not contradict with the mixed behavior policy which may not be Gaussian.
This is because there always exists a deterministic optimal policy \citep{puterman1990markov} and it is reasonable to model the deterministic policy with a uni-modal Gaussian distribution.
With fixed standard deviation $\sigma$, we can get rid of the first term and obtain
$$
\min_\phi \text{KL} \big( \pi_\phi(s)\big|\big|\hat{\pi}_b(s) \big) = \max_\phi \sum_{a\in\mathcal{A}} \pi_\phi(a|s) \log \hat{\pi}_b(a|s).
$$

Due to the complexity of behavior policies, we do not want to base the regularizer on an estimated $\hat{\pi}_b$.
Therefore, we want to formulate a sample-based behavior distribution $\hat{\pi}_b$.
For ease of notation, we assume the action space is discrete and the action samples do not overlap with each other.
Extending this formulation to continuous case is straightforward.
Given a state $s\in\mathcal{D}$, we define the probability mass function based on samples:
\begin{equation}
\hat{\pi}_b(a|s) = 
\begin{cases}
e^{-M+N} & \text{if } (s,a)\in\mathcal{D} \\
e^{-M} & \text{otherwise}
\end{cases},
\end{equation}
where $M>N>0$ and $M$ is a large constant so that $e^{-M}\to 0$, and they should satisfy the normalization condition $\sum_{a\in\mathcal{A}} \hat{\pi}_b(a|s) = 1$.

With this assumption, we have 
\begin{equation}
\begin{aligned}
\max_\phi & \sum_{a\in\mathcal{A}} \pi_\phi(a|s) \log \hat{\pi}_b(a|s) \\
= &
N \sum_{a: (s,a)\in\mathcal{D}} \pi_\phi(a|s) - M \sum_{a\in\mathcal{A}} \pi_\phi(a|s). \\
\end{aligned}
\end{equation}
The above optimization problem is equivalent to 
\begin{equation}
\label{eq:contrastive}
\max_\phi \left[
\sum_{a: (s,a)\in\mathcal{D}} \pi_\phi(a|s)
- \alpha
\sum_{a\in\mathcal{A}} \pi_\phi(a|s)
\right],
\end{equation}
for some hyperparameter $\alpha>0$. 
We can see that the first term is to maximize the probability of the action samples selected by the target policy (see also the red arrows in Fig. \ref{fig:reverse_KL}) and the second term is to minimize the probability on negative samples (see also the green arrows in Fig. \ref{fig:reverse_KL}) .
The combination of these two term can push the target policy away from selecting out-of-distribution actions.

Next, we replace $\pi_\phi(a|s)$ with the probability density function of Gaussian and obtain the following form of our regularizer:
\begin{equation}
\begin{aligned}
\max_\phi \Bigg[ &
\sum_{a: (s,a)\in\mathcal{D}} 
\exp\left(-\frac{(\mu_\phi(s)-a)^2}{2\sigma^2} \right)
\\
& - \alpha
\sum_{a\in\mathcal{A}} 
\exp\left(-\frac{(\mu_\phi(s)-a)^2}{2\sigma^2} \right)
\Bigg].
\end{aligned}
\end{equation}
To get rid of the effect of $\sigma$ that essentially serves only as a temperature hyperparameter, we remove the monotonically increasing exponential function and the coefficient $1/2\sigma^2$.
Moreover, to obtain negative action samples, we randomly sample two actions $a_1, a_2$ from the dataset $\mathcal{D}$ and use their mean $(a_1+a_2)/2$ as the negative sample.
In this way, the negative sample lies within the convex hull spanned by all the action samples.
At last, we obtain the regularizer that is applicable to the practical algorithm, i.e., updating the deterministic policy $\mu_\phi$ to minimize
$$
\mathbb{E}_{(s,a)\sim \mathcal{D}; a_1,a_2\sim\mathcal{D}}
\left[
\left(\mu_\phi(s) - a \right)^2 
-\alpha {\left(\mu_\phi(s) - \dfrac{a_1+a_2}{2} \right)^2}
\right].
$$

\begin{algorithm}[t]
\caption{TD3+RKL}
\label{algorithm:whole}
\begin{algorithmic}[1]
\State {\bf Input:} The offline dataset $\mathcal{D}$
\State {\bf Hyperparameters:}
Batch size $N$; Clip constant $c$; Standard deviation of target policy $\sigma$; Update frequency $d$
\State {Initialize the critic networks $Q_{\theta_1},Q_{\theta_2}$}
\State {Initialize the actor network $\mu_\phi$}
\State {Initialize the policy $\pi_\psi(\cdot|s) := \mathcal{N}(\mu_\psi(s), \sigma_\psi(s))$}
\State {$\triangleright$ \emph{Aleatoric Uncertainty Estimation}}
\State Estimate $\pi_\psi$ based on $\mathcal{D}$ to clone the behavior policy
\State $\hat{\beta}_b(\cdot) \leftarrow \beta_\phi(\cdot):= \log \sigma_\psi(\cdot)^2$
\State {$\triangleright$ \emph{Offline Reinforcement Learning}}
\For{$t=1,2,\cdots$}
\State {Sample $N$ transitions $\{(s_i,a_i,r_i,s_i')\}$ from $\mathcal D$}
\State {$a_i'\leftarrow \mu_{\bar{\phi}}(s_i')+\epsilon_i$; $\epsilon_i\sim\text{clip}(\mathcal{N}(0,\sigma))
,-c,c), \forall i \in[N]$}
\State {$y_i\leftarrow r_i+\gamma\min_{j=1,2}Q_{\bar\theta_j}(s_i',a_i'); \forall i \in[N]$}
\State {$\theta_j\leftarrow\mathop{\arg\min}_{\theta_i}\frac{1}{N}\sum_{i=1}^N (y_i-Q_{\theta_j}(s_i,a_i))^2; j=1,2$}
\If{$t$ mod $d = 0$}
\State {Construct $\lambda(s)$ using Eq. \eqref{eq:weights} and $\hat{\beta}_b(s)$}
\State {Update $\mu_\phi$ by optimizing Eq. (\ref{eq:TD3+RKL}) with $\lambda(s)$}
\State {Update target networks using Polyak averaging}
\EndIf
\EndFor
\end{algorithmic}
\end{algorithm}

\subsection{Adaptive Regularizer}
Since the variance of action samples changes on different states, we would better balance the RL and BC signals adaptively on different states according to the aleatoric uncertainty of the samples.
Our algorithm uses estimated standard deviation of the action samples conditioned on different states $\hat{\sigma}_b(s)$ as the aleatoric uncertainty.
Specifically, we use 
\begin{equation}
\label{eq:weights}
\lambda(s) = \frac{1}{1+\exp\left[ \zeta_1\hat{\beta}_b(s)-\zeta_2\right]} \in [0, 1],
\end{equation}
where $\hat{\beta}_b(s):=\log\hat{\sigma}_b(s)^2$ and $\zeta_1, \zeta_2$ are hyper-parameters.
The motivation for this formulation is to design numerically stable and well distributed weights with the sigmoid function and linear transformation.
In practical algorithms, we use the log-variance $\hat{\beta}_b(s)$ in a pre-estimated Gaussian policy $\pi_\psi$. 
\li{Notice that, different from the previous work that estimates a Gaussian policy to regularize the target policy, $\pi_\psi$ estimated in our algorithm is only used to adjust the weight and therefore is not required to be highly accurate.}

For the states on which $\hat{\sigma}_b(s)$ is large (i.e., the behavior policy takes a wide range of actions on these states), we reduce the BC regularization since we have sufficient knowledge on the effect of different actions and can select a good action among them to maximize the cumulative reward following the indication of the RL signal.
Otherwise, we have to restrict the policy to select only actions similar to the ones that have been tried by the behavior policy.

\subsection{The TD3+RKL Algorithm}

We present the \li{details} 
of TD3+RKL in Algorithm \ref{algorithm:whole}. 
Note that we also adopt the useful tricks used in TD3+BC (such as using target networks, pre-normalizing the states in the dataset, and adaptively adjusting the weight of the RL signal by dividing $\mathbb{E}_{(s,a)\sim\mathcal{D}}[Q(s,a)]$) and do not present them in the algorithm block for simplicity.

In Line 6-8, we estimate a behavior policy that is modeled as a Gaussian distribution conditioned on $s\in\mathcal{S}$ parameterized by $\mu_\psi:\mathcal{S}\to\mathbb{R}$ and $\beta_\psi:\mathcal{S}\to\mathbb{R}$ with $\beta_\psi(s) := \log \sigma_\psi(s)^2$.
\li{This process is similar to the behavior cloning process in many offline algorithms such as \citep{kumar2020conservative}.
Although this cloned behavior policy may not accurately approximate the mixed behavior policy used for dataset collection, we only use the log-variance of the cloned policy which does not require a high accuracy to calculate the adaptive weights for later offline RL process.}
Later in Section \ref{sec:learned_aleatoric}, we will show that the estimated variance can nicely reveal the aleatoric uncertainty and result in reasonable weights.

In Line 9-20, we learn the Q function following the TD3 algorithm \citep{fujimoto2018addressing,fakoor2021continuous}.
The key of our algorithm is to calculate the adaptive weights for each state in the batch following Eq.~\eqref{eq:weights} (cf. Line 16) and  update the target policy by optimizing both the RL and BC signals following the update rule defined in Eq.~\eqref{eq:TD3+RKL} (cf. Line 17).
\section{Experiments}
\label{sec:experiments}

In this section, we conduct experiments to evaluate our algorithm from the following aspects\footnote{Codes are available at https://github.com/yuanying-cc/TD3-RKL.}:
\begin{itemize}
    \item We compare TD3+RKL with the previous offline RL algorithms on the standard D4RL datasets as well as several new mixed datasets to evaluate the performance of different algorithms on the datasets collected by a mixture of policies.
    \item For the adaptive weights, we study how well can we learn the aleatoric uncertainty $\hat{\sigma}_b$ on both a toy example and the Halfcheetah task with D4RL datasets~\citep{fu2020d4rl}.
    \item For the new behavior cloning regularizer, we further conduct controlled experiments to compare the performance of different behavior cloning regularizers on the MuJoCo locomotion tasks using the D4RL datasets.
\end{itemize}

In our experiments, we choose $\zeta_1=10$ and $\zeta_2=5$ according to pre-estimated log-variance on the action samples. 
For simplicity, we use $\alpha=1.0$ which results in reasonable performance. 
Other hyper-parameters are set following the implementation of TD3+BC.
We use the open-source D4RL datasets in our experiments and will release our code when the paper is published.

\begin{table*}[htb]
\caption{\normalfont Comparison of difference algorithms on Mujoco tasks using the D4RL dataset and mixed datasets.
The first four groups of benchmarks use the standard D4RL datasets \cite{fu2020d4rl}, and the last three groups use new datasets generated by mixing the standard datasets.
The scores for BC, BRAC, and TD3+BC for the first four groups are taken from the original papers, while the other scores are obtained by our implementation and averaged over 10 evaluations with 5 seeds.
}
\label{tab:comparison}
\centering
\begin{tabular}{lrrrrrl}
\toprule
    
{\bf Task Name} & {\bf BC} & {\bf BRAC} & {\bf CDC} & {\bf TD3+BC} &  {\bf TD3+RKL}  \\
\midrule
{HalfCheetah-Random}  & 2.0$\pm$0.1 & 23.5 & {\bf27.4} & 10.2$\pm$1.3 &  23.2$\pm$1.1  \\
{Hopper-Random}   & 9.5$\pm$0.1 & 11.1 & {\bf14.8} & 11.0$\pm$0.1 & 11.1$\pm$0.1   \\
{Walker2d-Random}  & 1.2$\pm$0.2 & 0.8 & {\bf7.2} & 1.4$\pm$1.6 & 2.2$\pm$1.3 &   \\
\midrule
{HalfCheetah-MediumReplay}  & 37.4$\pm$1.8 & 45.6 & 44.7 & 43.3$\pm$0.5 & {\bf47.1}$\pm$1.5  \\
{Hopper-MediumReplay} & 19.7$\pm$5.9 & 0.7& {\bf55.9} & 31.4$\pm$3.0 & 45.3$\pm$2.5&  \\
{Walker2d-MediumReplay}   & 8.3$\pm$1.5 & -0.3 & 23.0 & {\bf25.2$\pm$5.1} & 24.5$\pm$2.3 &   \\
\midrule
{HalfCheetah-MediumExpert}  & 67.6$\pm$13.2 & 43.8 & 59.6 &  97.9$\pm$4.4 & {\bf105.4}$\pm$4.8\\
{Hopper-MediumExpert}  & 89.6$\pm$27.6 & 1.1& 86.9 & {\bf111.2}$\pm$0.2 & 112.0$\pm$ 0.6  \\
{Walker2d-MediumExpert}  & 12.0$\pm$5.8 & -0.3 & 70.9& {\bf101.1}$\pm$9.3 &  98.4$\pm$10.6&   \\
\midrule
{HalfCheetah-Expert}  & 105.2$\pm$1.7 & 3.8& 82.1 & 105.7$\pm$1.9 & {\bf106.4}$\pm$2.9\\
{Hopper-Expert}   & 111.5$\pm$1.3 & 6.6 & 102.8 & 111.2$\pm$0.2 & {\bf112.3}$\pm$0.6  \\
{Walker2d-Expert}   & 56.0$\pm$24.9 & -0.2& 87.5 & 105.7$\pm$2.7 & {\bf106.9}$\pm$ 1.6  \\
\midrule
Average on D4RL datasets & 43.3 & 11.4 & 55.2 & 62.9 & {\bf66.2} & (+5.2\%) \\
\midrule
{HalfCheetah-Random-MediumReplay}  &24.2$\pm$5.0 & 36.0$\pm$4.2 & 28.9$\pm$5.2& 33.6$\pm$1.9 & {\bf40.9}$\pm$4.8 \\
{Hopper-Random-MediumReplay}  &11.4$\pm$8.3 & 8.2$\pm$3.3 & 26.0$\pm$7.9& 19.0$\pm$2.4 & {\bf28.3}$\pm$5.2 \\
{Walker2d-Random-MediumReplay}  &3.3$\pm$2.6 & 0.6$\pm$0.2 & 13.7$\pm$6.8 & 5.8$\pm$2.6 & {\bf16.7}$\pm$5.9 \\
\midrule
{HalfCheetah-Random-MediumExpert}  &43.9$\pm$19.6 & 37.7$\pm$6.8 & 34.2$\pm$7.7 & 84.5$\pm$6.5 & {\bf101.1}$\pm$8.6\\
{Hopper-Random-MediumExpert}  &41.2$\pm$22.7 & 6.2$\pm$2.5 & 34.4$\pm$3.7& 103.2$\pm$2.8 & {\bf106.5}$\pm$2.5 \\
{Walker2d-Random-MediumExpert}  &5.6$\pm$2.9 & 1.0$\pm$0.2 & 7.9$\pm$1.3 & 5.4$\pm$1.1 & {\bf24.9}$\pm$7.1 \\
\midrule
{HalfCheetah-Random-Expert}  &47.3$\pm$17.2 & 16.9$\pm$3.7 & 38.5$\pm$8.3 & 89.8$\pm$4.4 & {\bf107.5}$\pm$5.2\\
{Hopper-Random-Expert}  &69.6$\pm$24.4 & 9.4$\pm$1.5 & 88.4$\pm$9.6& 99.7$\pm$10.5 & {\bf111.8}$\pm$4.5 \\
{Walker2d-Random-Expert}  &11.3$\pm$18.8 & 0.5$\pm$0.1 & 8.6$\pm$2.1 & 3.5$\pm$0.2 & {\bf20.7}$\pm$6.4 \\
\midrule
Average on mixed datasets & 28.6 & 12.9 & 31.2 & 49.4 & {\bf62.0} & (+25.5\%) \\
\bottomrule
\end{tabular}
\end{table*}
\begin{figure}[t]
   \centering
   \includegraphics[width=0.8\columnwidth]{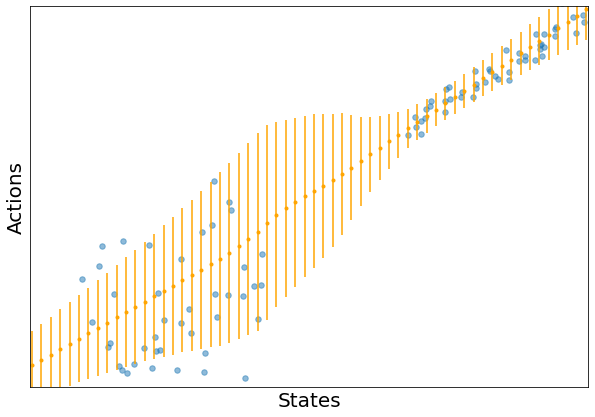}
   \caption{
   Evaluation of the learned aleatoric uncertainty on a toy example with one-dimensional states and actions. 
   The x-axis represents the state and the y-axis represents the action.
   The orange bars represent the predicted aleatoric uncertainty on different states. 
   }
\label{fig:toy_example}
\end{figure}
\begin{figure}[tb]
   \centering
   \includegraphics[width=0.8\columnwidth]{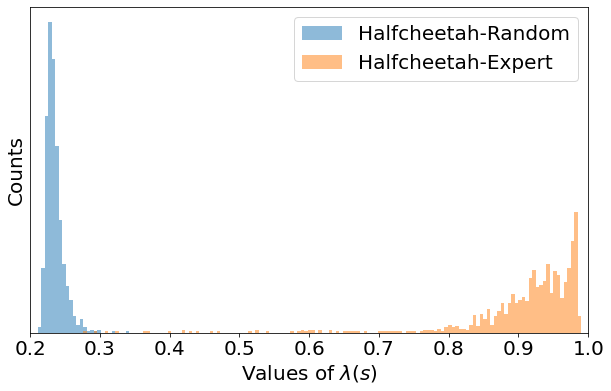}
   \caption{
    Histogram of per-state BC weights $\lambda(s)$ computed with the learned aleatoric uncertainty of the behavior policy in the Halfcheetah-random and Halfcheetah-expert D4RL datasets. 
    The blue (orange) bars represent the uncertainty learned from the random (expert) dataset. 
    The x-axis represents the weight value and the y-axis represents the count.
   }
\label{fig:toy_example_halfcheetah}
\end{figure}
\begin{figure*}[tb]
   \centering
   \includegraphics[width=1.8\columnwidth]{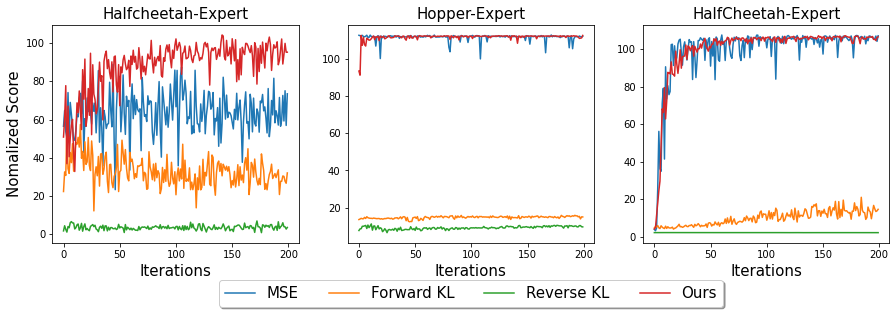}
   \caption{
  Comparison of the performance learned only using BC signals on expert datasets from D4RL. We use four kinds of BC signals listed in Table 1. The reverse and forward KL signals are used as constraints in BRAC and CDC. Each score is averaged over 10 evaluations with 5 random seeds. The x-axis represents the evaluation time steps while the y-axis represents the normalized scores. 
   }
\label{fig:bc_signals}
\end{figure*}

\subsection{Comparison with Previous Offline RL Algorithms}

In this part, we compare TD3+RKL with existing policy-constraint based offline RL algorithms.
To evaluate the performance of different algorithms when learning from mixed datasets, we obtain new datasets by mixing the random D4RL dataset with the others, resulting in Random-MeduimReplay, Random-MeduimExpert, and Random-Expert datasets.
We evaluate the algorithms on standard D4RL datasets and the new mixed datasets. 
We present the result in Table~\ref{tab:comparison}.
First, we observe that existing methods perform poorly on the new mixed datasets.
Notice that, even if the sizes of the mixed datasets are larger than those of the D4RL datasets (since they are the combination of two of the D4RL datasets), the algorithms perform worse on the mixed datasets.
This indicates that learning from mixed datasets is a harder task than learning from datasets generated by pure policies.
Second, we can observe that TD3+RKL outperforms the previous algorithms on most D4RL datasets, resulting in an average performance increase of 5.3\% compared with the best baseline TD3+BC.
Third, we can see that TD3+RKL outperforms the other baselines significantly on the new mixed datasets and achieves an average normalized score that is 25.5\% higher than TD3+BC.
This suggests that TD3+RKL performs well on the scenarios where the dataset is collected by a mixture of distinct policies.

\subsection{Evaluation on Learned Aleatoric Uncertainty}
\label{sec:learned_aleatoric}

Recall that, we estimate the aleatoric uncertainty conditioned on different states and adaptively balance the RL and BC signals based on the estimated uncertainty in TD3+RKL.
Therefore, the effectiveness of adaptive weights depends on the quality of the estimated uncertainty.
In this part, we evaluate the quality of the learned aleatoric uncertainty and show the resultant weights. 

\textbf{Evaluation on the toy example.} 
We first evaluate the aleatoric uncertainty extracted from the cloned policy $\pi_\psi$ modeled using Gaussian on a toy example.
In this toy example, we consider the one-dimensional state and action spaces.
We show the state-action pairs in the training dataset with blue points in Figure \ref{fig:toy_example}, where the x-axis represents the state and the y-axis represents the action.
Based on the dataset, we learn the aleatoric uncertainty as in TD3+RKL and show the uncertainty of the action samples with the orange bars in Figure \ref{fig:toy_example}, the lengths of which represent the level of uncertainty. 

We can observe that the action samples on the left have higher uncertainty than those on the right.
Accordingly, the predicted uncertainty on the left is larger than that on the right, which indicates that the learned aleatoric uncertainty can nicely capture the uncertainty or coverage of the action samples on different states.

\textbf{Evaluation on the Halfcheetah dataset from D4RL.} 
To further evaluate the effectiveness of uncertainty estimation combined with our weight formulation on robotic control tasks, we present the per-state BC weight $\lambda(s)$ on the Halfcheetah-Random and Halfcheetah-Expert dataset from D4RL. 
We show the weight $\lambda(s)$ on the states from different datasets with a histogram in Figure \ref{fig:toy_example_halfcheetah}.
The weight is calculated based on the aleatoric uncertainty learned on the combination of both datasets.
The orange bars represent the weights for the samples from the expert dataset while the blue bars represent the weights for the samples from the random dataset. 
We expect that the random dataset has larger action coverage and aleatoric uncertainty than the expert dataset and therefore smaller BC weights.
We can see that our experiment result is consistent with this intuition, which indicates the effectiveness of the adaptive weight formulation on standard offline datasets.

\subsection{Effects of Different BC Signals}

In this part, we design controlled experiments to further compare different kinds of behavior cloning signals listed in Table \ref{tab:comp_rule}.
We evaluate the per-state weighted reverse KL divergence used in TD3+RKL (ours), the MSE loss used in TD3+BC \citep{fujimoto2021minimalist}, the forward KL used in CDC \citep{fakoor2021continuous} and the reverse KL used in BRAC \citep{wu2019behavior}. 
The first two BC signals are used for learning deterministic policies and the latter two BC signals are used for learning stochastic policies.
To study the effect of pure BC signals, we evaluate them under the behavior cloning setting, i.e., we do not use any RL signal.
We use the expert datasets from the D4RL datasets in these experiments. 

We show the experiment results in Figure \ref{fig:bc_signals}. 
First, we note that although BRAC also uses the reverse KL divergence, its performance in pure behavior cloning is poor.
This may result from the implicit entropy maximization for the target policy which is not suitable especially for the expert dataset.
Second, we observe that
the MSE loss used in TD3+BC works significantly better than the losses in BRAC and CDC that optimize the stochastic target policy.
This motivates us to learn a deterministic target policy that is simple as well as effective. 
At last, the per-state weighted loss derived from minimizing the reverse KL divergence used in our algorithm learns a better policy than TD3+BC on the Walker2d task and achieves comparable performance on the other two tasks.
This shows that our BC regularization can clone the behavior policy well given an expert dataset.
\section{Conclusion}
\label{sec:conclusion}

In this paper, we consider the offline reinforcement learning (RL) setting where the agent should learn from a dataset collected by a mixture of behavior policies.
For this setting, the algorithm should not only balance the RL and behavior cloning (BC) signals adaptively on different states but also avoid selecting out-of-distribution actions in the face of multi-modal behavior action distributions.
\li{To meet these requirements, we propose TD3+RKL (reverse KL divergence) that uses the BC regularzier derived from  adaptively weighted reverse KL divergence with a deterministic target policy. }
Our method is simple since it does not require us to estimate an accurate behavior policy or maintain a stochastic target policy. 
Nevertheless, our method is effective empirically when learning from both the standard D4RL datasets and the new dataset generated by mixing the samples collected by different behavior policies.


We note that, although our algorithm outperforms the previous methods on mixed datasets, the performance of the policy learned based on a mixed dataset (e.g., Walker2d-Random-Expert) is not as good as that based on a subset of this dataset (e.g., Walker2d-Expert).
This motivates us to adjust the behavior cloning regularizer based on not only the variance of the action samples conditioned on the state but also the performance of the underlying behavior policy. 
However, this requires more complicated techniques to discriminate and evaluate the underlying behavior policies.
We leave it as a future research direction.

\bibliographystyle{IEEEtranN}
\bibliography{main}

\end{document}